\documentclass{article}

\usepackage{spconf,amsmath,graphicx}
\usepackage{bm}
\usepackage{amsfonts}
\usepackage{booktabs}
\usepackage{pifont}
\usepackage{subfigure}
\usepackage{algorithm}
\usepackage{multicol}
\usepackage{lineno,hyperref}
\usepackage{multirow}
\usepackage{algpseudocode}
\usepackage{tabularx}
\usepackage{verbatim}
\usepackage{longtable}
\usepackage{makecell}
\usepackage{threeparttable}

\newcommand{\MCols}[2]{\multicolumn{#1}{c|}{#2}}
\usepackage[nameinlink]{cleveref}
\Crefformat{equation}{#2Eq.~(#1)#3}
\title{DMFORMER: CLOSING THE GAP BETWEEN CNN AND VISION TRANSFORMERS}

\def\ouratt{DMA}
\def\ourmethod{DMFormer}
\name{Zimian Wei$^{1}$,  Hengyue Pan$^{1}$, Lujun Li$^{2}$, Menglong Lu$^{1}$, Xin Niu$^{1}$, Peijie Dong$^{1}$, Dongsheng Li$^{1}$}
\address{$^1$ College of Computer, National University of Defense Technology\\
 $^2$ Chinese Academy of Sciences, Beijing, China\\
 }

\begin{document}
%\ninept
%
\maketitle
\begin{abstract}

% Vision transformers have shown excellent performance in computer vision tasks.
% However, the computation cost of their self-attention mechanism is expensive.
% % Comparatively, CNN is more efficient with built-in inductive bias.
% Recent works replace the self-attention mechanism in vision transformers with convolutional operations, which is more efficient with built-in inductive bias.
% % can achieve comparable results.
% Nevertheless, they either ignore multi-level features or lack dynamic prosperity, leading to sub-optimal performance.
% In this paper, we propose a Dynamic Multi-level Attention mechanism (\ouratt{}), which captures different patterns of input images by multiple kernel sizes and enables input-adaptive weights with a gating mechanism.
% Based on \ouratt{}, we present an efficient backbone network named \ourmethod{}.
% \ourmethod{} adopts the overall architecture of vision transformers, while replacing the self-attention mechanism with our proposed \ouratt{}.
% Extensive experimental results on ImageNet-1K and ADE20K datasets  demonstrated that \ourmethod{} achieves state-of-the-art performance, which outperforms similar-sized vision transformers(ViTs) and convolutional neural networks (CNNs).

Vision transformers have shown excellent performance in computer vision tasks. 
As the computation cost of their self-attention mechanism is expensive, recent works tried to replace the self-attention mechanism in vision transformers with convolutional operations, which is more efficient with built-in inductive bias. However, these efforts either ignore multi-level features or lack dynamic prosperity, leading to sub-optimal performance.
In this paper, we propose a Dynamic Multi-level Attention mechanism (\ouratt{}), which captures different patterns of input images by multiple kernel sizes and enables input-adaptive weights with a gating mechanism.
Based on \ouratt{}, we present an efficient backbone network named \ourmethod{}.
\ourmethod{} adopts the overall architecture of vision transformers, while replacing the self-attention mechanism with our proposed \ouratt{}.
Extensive experimental results on ImageNet-1K and ADE20K datasets  demonstrated that \ourmethod{} achieves state-of-the-art performance, which outperforms similar-sized vision transformers(ViTs) and convolutional neural networks (CNNs).

\end{abstract}

\section{Introduction}

Recently, vision transformers (ViT) \cite{vit, deit} have drawn growing attention in computer vision research.
Due to the capacity of modeling long-range dependencies, ViTs are specialized in extracting global features. 
However, the self-attention mechanism in transformers brings about heavy computation costs, making them unaffordable for high-resolution downstream tasks(e.g., semantic segmentation). 
Although recent local vision transformer methods \cite{swin, huang2021shuffle} alleviate this problem to some extent, the implementation of cross-window strategies in local self-attention is still sophisticated. 

As a complementary, convolution neural network(CNN) focus on capturing local relations with high efficiency. 
With built-in inductive biases, CNNs are easy to train with quick convergence. 
To this end, there is a trend to take the merits of both CNNs and ViTs by migrating desired properties of ViTs to CNNs, including the overall architecture design, large receptive field, and data specificity provided by the attention mechanism.
For example, ConvNext \cite{liu2022convnet} built a pure CNN family based on ResNet \cite{resnet}, which performs on par or slightly better than ViT by learning their training procedure and macro/micro-level architecture designs.
RepLKNet \cite{ding2022scaling} adopts as large as $31\times 31$ kernel size to enlarge effective receptive fields following the design in ViT.
Although encouraging performance has been achieved by the above methods, their computation costs are relatively large.
\cite{han2021connection} competes favorably with Swin transformer \cite{swin} by replacing the local self-attention layer with the dynamic depth-wise convolution layer, while keeping the overall structure unchanged.
However, the lack of multi-level features limits its capacity to achieve better performance.

% altered the landscape of network architecture design with powerful performance and nice scalability.

\begin{figure}[t]
  \centering
    \includegraphics[width=\linewidth]{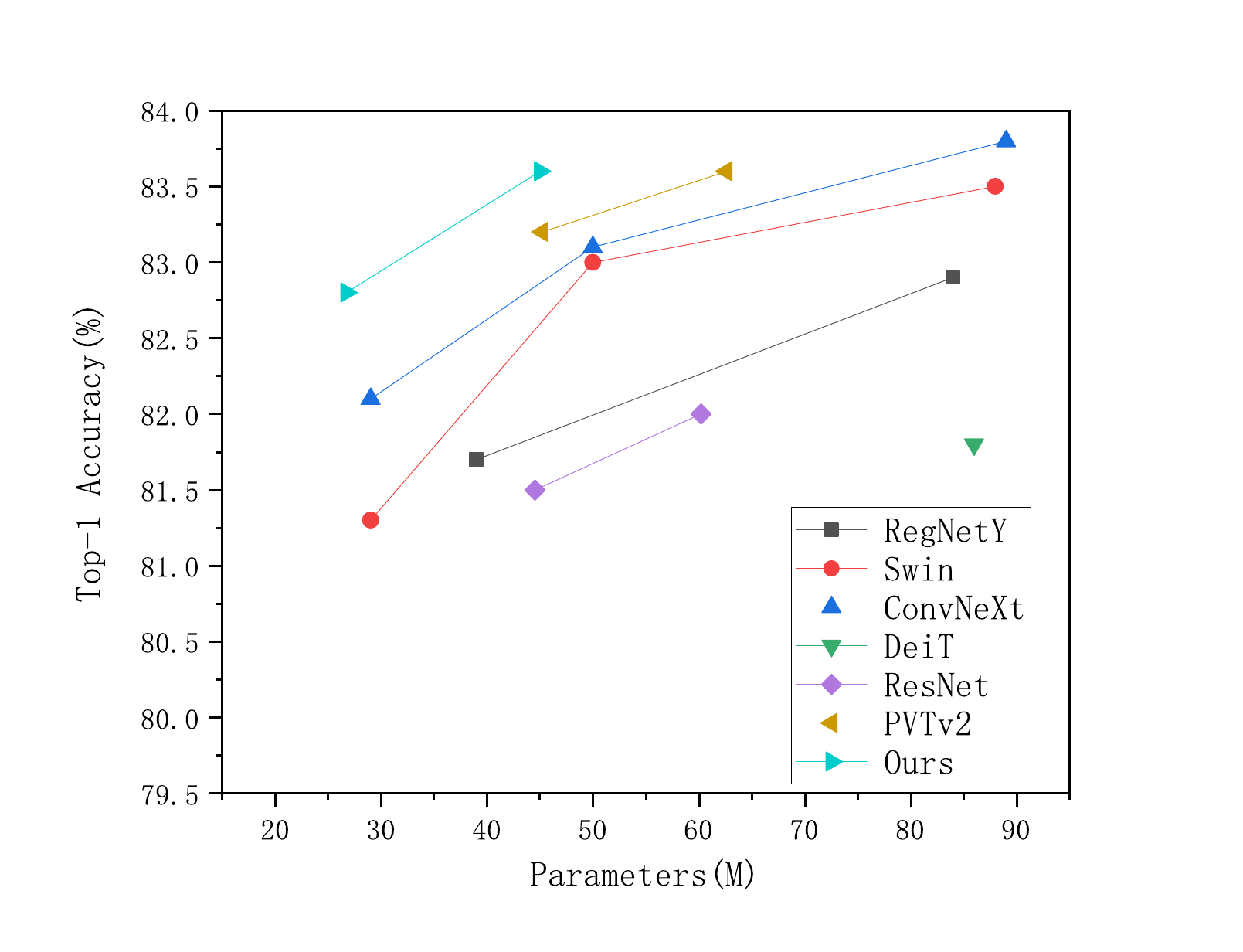}
  \vspace{-10mm}
   \caption{
    Results of different models on ImageNet-1K validation set.
We compare accuracy-parameters trade-off of recent models RegNet~\cite{regnet}, Swin Transformer~\cite{swin}, ConvNeXt~\cite{liu2022convnet}, DeiT~\cite{deit}, ResNet~\cite{resnet}, PVTv2 \cite{wang2021pvtv2} and our \ourmethod{}.
% Above: Accuracy-Parameters trade-off diagram..
%   Visualization results produced by using Grad-CAM~\cite{selvaraju2017grad} on images selected from the ImageNet validation dataset. 
% % 	All images come from different categories in ImageNet validation set. CAM is produced by using Grad-CAM~\cite{selvaraju2017grad}. 
% 	We present different CAMs generated by  Swin-T~\cite{swin}, ResNet50~\cite{resnet} and ours.
	}
   \label{fig:performance_tradeoff}
   \vspace{-4mm}
\end{figure}

\begin{table}[t]\centering
	\caption{Favorable properties correspond to different modules, including self-attention, multi-scale convolution, dilated convolution, and \ouratt{}.}
\footnotesize
	\vspace{2mm}
	\resizebox{\linewidth}{!}{
\begin{tabular}{ccccc}
\hline
Properties              & \begin{tabular}[c]{@{}c@{}}Self-\\ attention\end{tabular} & \begin{tabular}[c]{@{}c@{}}Multi-scale \\ Convolution\end{tabular} & \begin{tabular}[c]{@{}c@{}}Dilated \\ Convolution\end{tabular} & \ouratt{} \\ \hline
Local inductive bias    &    \ding{55}         &  \ding{51}                                                                 &  \ding{51}                                                               &  \ding{51}   \\
Large receptive field   &  \ding{51}                                                         & \ding{55}                                                                   &   \ding{51}                                                             &  \ding{51}   \\
Multi-scale Interaction   &   \ding{55}                                                        &  \ding{51}                                                                  &  \ding{51}                                                              &  \ding{51}    \\
Input Adaptive          &   \ding{51}                                                        &  \ding{55}                                                                  &    \ding{55}                                                            &  \ding{51}   \\
% Computation Complexity  &     $\mathcal{O}\left( n^2 \right)$                                                      &  $\mathcal{O}\left( n \right)$                                                                  &     $\mathcal{O}\left( n \right)$                                                           &  $\mathcal{O}\left( n \right)$   \\ 
\hline
\end{tabular}}
	\label{tab:property}
	  	\vspace{-5mm}
\end{table}

% including the overall architecture design, large receptive filed, and data specificity provided by the attention mechanism.

\begin{figure*}[t]
  \centering
    \includegraphics[width=0.9\linewidth]{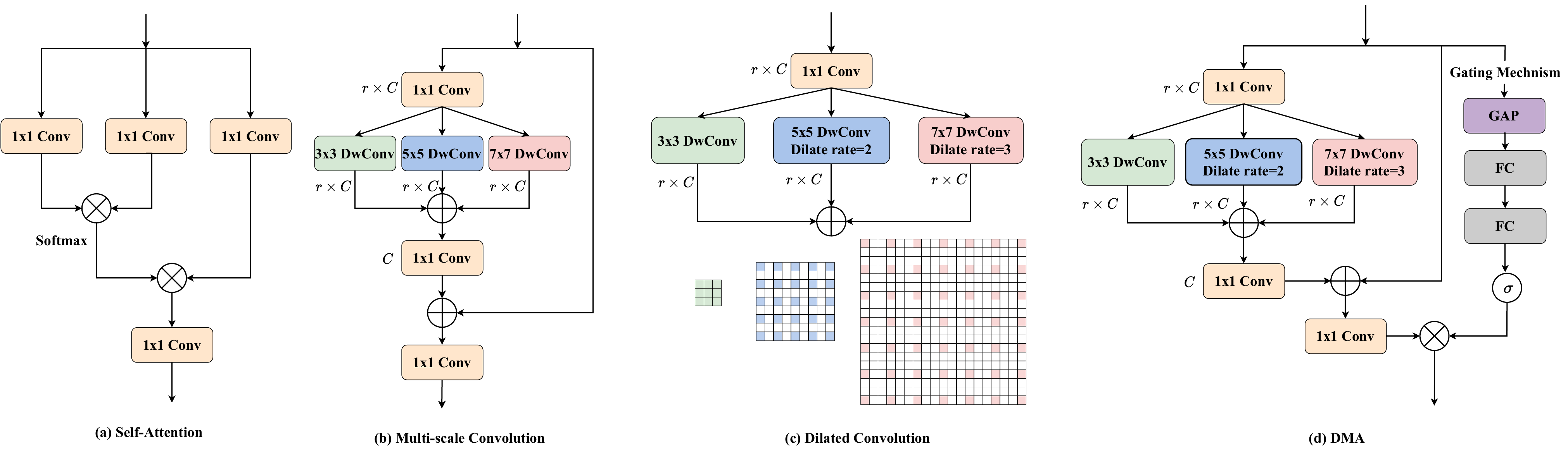}

  \caption{The structure comparison between related modules: self-attention (a), multi-scale convolution (b), dilated convolution (c), and \ouratt{} (d).
  DwConv, GAP, $\sigma$, and FC refer to depth-wise convolution, global average pool, sigmoid function, and fully connected layer, respectively. 
%   Dilation a and b are the dilation rates in the depth-wise convolutions.
  $C$ refers to the number of channels, while $r$ is the expansion ratio in the \ouratt{}.
    % (d) The properties corresponding to different modules.
%   Following \cite{yu2021metaformer, swin}, \ourmethod{} apply the hierarchical architecture design with 4 stages, and each stage consists of multiple blocks. $C_i, N_i$ refer to the feature dimension and block number in stage $i$, respectively.
%   The basic block in \ourmethod{}. We adopt the general architecture of blocks in vision transformer, while replace the 
%   (local) self-attention layer with \ouratt{}.
  }
  \label{fig:MCA}
  \vspace{-5mm}
\end{figure*}

In this paper, we propose a new dynamic multi-level attention (\ouratt{}) mechanism (see Table \ref{tab:property} and Fig.~\ref{fig:MCA} for more details).
% The advantages of \ouratt{} are three folds.
Firstly, \ouratt{} is characterized by applying multiple kernel sizes for different resolution patterns. 
As mentioned in \cite{tan2019mixconv}, multiple kernel sizes can effectively improve the model's performance. 
Secondly, \ouratt{} involves dilated convolutions to efficiently enlarge the receptive field.
% provide input adaptability, with which the channel-wise relationships are captured and the representational power of the network is enhanced. 
Thirdly, we design a lightweight gating mechanism in \ouratt{} to provide input adaptability, with which the channel-wise relationships are captured and the representational power of the network is enhanced. 
% The channel-wise relationships are captured and the representational power of the network is enhanced. 
% The information from different channels is incorporated, and the representational power is enhanced.
% The information from different channels is incorporated, which effectively enhances the representational power. 
Based on \ouratt{}, we extend the architecture of Swin Transformer  \cite{swin} and propose a new framework named \ourmethod{}. 
The experimental results show that \ourmethod{} achieves state-of-the-art performance on ImageNet classification (see Fig. \ref{fig:performance_tradeoff}) and semantic segmentation tasks.

In summary, the contributions of this paper have two aspects:
1) We propose a new attention mechanism named \ouratt{}, which combines the advantages
of convolution and self-attention.
% including efficiency, input adaptive, multi-level interaction, and large receptive field.
% is dynamic and adopts multiple convolution kernel sizes for different resolution patterns. 
2) Based on \ouratt{}, we design a backbone model named \ourmethod{}.
Extensive experiments on ImageNet classification and semantic segmentation tasks show the superior of \ourmethod{} over other existing models.

\section{Approaches}
% In this section, we depict details of the proposed cross-modal distillation method 
In this section, we first present details of \ouratt{}.
% % each component in the overall structure illustrated in Figure \ref{fig:overall_architecture}, including , MLP, and Conv Stem.
% % the  \ouratt{} in Section \ref{}.
Then we show the overall structure and present different architecture designs in \ourmethod{} family.
% In the first stage, we mutually train the textual and visual networks and improve the capacity of each network by a novel cross-modal knowledge distillation method. Then in the second stage, we fix the well-trained textual and visual networks and train a fusion mechanism to achieve better performance for multi-modal fake news detection. 

\subsection{\ouratt{}}
\label{Sec:MCA}
% The main idea of \ouratt{} is that 
% It is demonstrated in \cite{} that both high
We illustrate the structure of \ouratt{} in Fig. \ref{fig:MCA} (d).
The two key components in \ouratt{} are the multi-scale dilated convolution and gating mechanism. 
% Multi-scale dilated convolution is able to combining multi-scale feature maps
Multi-scale dilated convolution captures different patterns with various resolutions of input images, while the gating mechanism learns to selectively emphasize informative features by re-calibration.
% Meanwhile, the gating mechanism learns to selectively emphasize informative features by feature re-calibration.
% which learns to selectively emphasize informative features and suppress nontrivial ones.
% Meanwhile, the gating mechanism is introduced to perform feature re-calibration, which learns to selectively emphasize informative features and suppress nontrivial ones.
% To obtain multi-level features 

Assuming that the input of \ouratt{} is $X$.
As depicted in Fig. \ref{fig:MCA} (d),  a $1\times 1$ convolution layer ($Conv_{\exp \_r}1\times 1$) is applied to expand the number of channels by $r$ times.
Then, a parallel design of $3\times 3$, $5 \times5$, $7\times 7$ depth-wise convolution is introduced to learn multi-scale features.
BatchNorm and ReLU are followed to prevent over-fit when training.
Next, in order to apply a residual connection for better optimization, we apply a $1\times 1$ convolution layer to reduce the number of channels as the original input $X$.
The above operators can be expressed as follows:
\begin{small}
\begin{align}
\small
X_E & =Conv_{\exp \_r}1\times 1\left( X \right), 
\notag \\
X_1,X_2,& X_3  =Parallel_{3\times 3,5\times 5,7\times 7}\left( X_E \right), 
\notag \\
X_P & =\mathrm{Re}LU\left( BN\left( X_1+X_2+X_3 \right) \right), 
\notag \\
X^{\prime} & = X+Conv1\times 1\left( X_P \right),
\end{align}
\end{small}
where $Parallel_{3\times 3,5\times 5,7\times 7}$ contains multi-branch of $3\times 3$,~$5\times 5$,~$7\times 7$ convolution layers. 
% Following \cite{ding2022scaling, guo2022visual}, 
Specifically, for branches with kernel size $5\times 5$ and ~$7\times 7$, we set the corresponding dilated rates as $2,3$ to obtain larger receptive field. 
% Since the MetaFormer block already has a residual connection, subtraction of the input itself is added in Equation (\ref{eq:pool}). The PyTorch-like code of the pooling is shown in Algorithm \ref{alg:code}.

For the gating mechanism, we apply a global average pooling (GAP) layer
to obtain global information, followed by two successive fully connected layers. 
At last, a sigmoid function is applied to compute the attention vector.
The operations in the gating mechanism can be formulated as follows:
\begin{small}
\begin{align}
    &V =\mathrm{ReLU}  \left( FC\left( GAP\left( X \right) \right) \right),
 \notag \\
& Attn=   Sigmoid \left( FC\left( V \right) \right),
\end{align}
\end{small}

Finally, the output of \ouratt{} is obtained by re-calibrating the fused feature $X^{\prime}$ with the gating mechanism as follows:
\begin{small}
\begin{equation}
    Output=Attn\otimes Conv1\times 1\left( X^{\prime} \right),
\end{equation}
\end{small}
where $\otimes$ means the element-wise matrix multiplication.

In the following section, we will introduce the basic block, overall framework, and detailed architectural design in \ourmethod{} family.

\subsection{\ourmethod{}}
\label{Sec:ConvFomer}

\begin{figure}[ht]
  \centering
    \includegraphics[width=\linewidth]{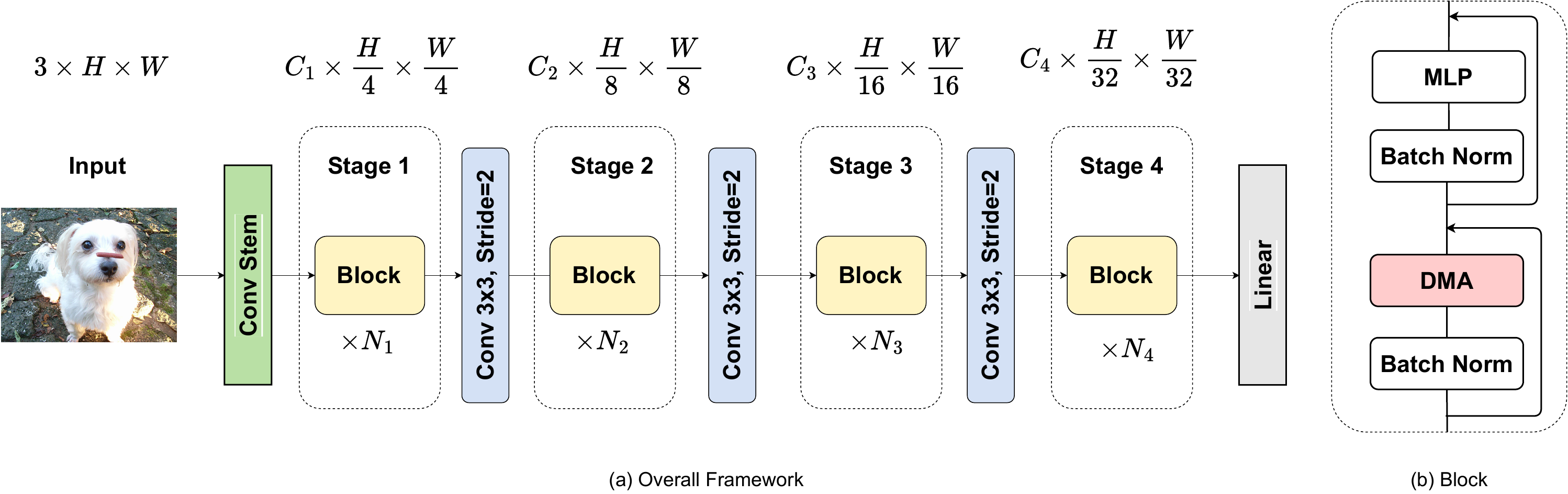}
%   \vspace{-5mm}
  \caption{(a) The overall framework of \ourmethod{}. Following \cite{swin}, \ourmethod{} adopts the hierarchical architecture with 4 stages, and each stage consists of multiple blocks. $C_i, N_i$ refer to the feature dimension and block number in stage $i$, respectively.
    (b) The basic block in \ourmethod{}. We apply the modular design in vision transformers, while replacing the 
    self-attention layer with \ouratt{}.}
  \label{fig:overall_architecture}
  \vspace{-4mm}
\end{figure}

We build \ourmethod{} with a hierarchical design similar to traditional CNNs \cite{resnet} and recent local vision transformers \cite{swin}. 
Fig. \ref{fig:overall_architecture} (a) presents the overall framework of \ourmethod{} and Fig. \ref{fig:overall_architecture} (b) shows the basic block in \ourmethod{}.

\begin{table}[t]
  \centering
  \footnotesize
  \setlength{\tabcolsep}{3.pt}
  \footnotesize
  \caption{The detailed setting for different versions of \ourmethod{}.
  ER, r represent the expansion ratio in the MLP module and \ouratt{} module, respectively.
%   r is the expansion ratio in the \ouratt{} module.
  }\label{tab.architecture}
  	\vspace{2mm}
  \begin{tabular}{c|c|c|c|c|c}
    \toprule 
    stage & output size & ER & r & \ourmethod{}-S & \ourmethod{}-L   \\ \hline
    1 & $\frac{H}{4}\times \frac{W}{4} \times C_1$ & 8 & 4 & 
    \makecell{$C_1=64$ \\ $N_1=2$} & \makecell{$C_1=64$ \\ $N_1=3$}   \\ \hline
    2 & $\frac{H}{8}\times \frac{W}{8} \times C_2$ & 8  & 4 &
    \makecell{$C_2=128$ \\ $N_2=2$} & \makecell{$C_2=128$ \\ $N_2=3$}   \\ \hline
    3 & $\frac{H}{16}\times \frac{W}{16} \times C_3$ & 4  & 4 & 
    \makecell{$C_3=320$ \\ $N_3=6$} & \makecell{$C_3=320$ \\ $N_3=12$}   \\ \hline
    4 & $\frac{H}{32}\times \frac{W}{32} \times C_4$ & 4  & 4 &
    \makecell{$C_4=512$ \\ $N_4=2$} & \makecell{$C_4=512$ \\ $N_4=3$}  \\ \hline
    \MCols{4}{Parameters (M)} & 26.7 & 45.0 \\ \hline
    \MCols{4}{{FLOPs} (G)}    & 5.0 & 8.7  \\ 
    \bottomrule
  \end{tabular}
  	\label{tab:model}
  	 % 	\vspace{-4mm}
\end{table}

The input image $I$ is first processed by the convolution stem module, which consists of a $7\times7$ convolution layer with a stride of 2, a $3\times 3 $ convolution layer with a stride of 1, and a non-overlapping $2\times 2$ convolution layer with a stride of 2.
Then the spatial size of output features after the convolution stem module is $\frac{H}{4}\times \frac{W}{4}$.

% As shown in Figure \ref{fig:overall_architecture}, the overall framework of \ourmethod{} is constructed by basic blocks.
% where the token mixer is not specified while the other components are kept the same as transformers. 
% Within one block,  such as  patch embedding for ViTs \cite{vit},
\begin{small}
\begin{equation}
    X = \mathrm{ConvStem}(I),
\end{equation}
\end{small}
where $X\in \mathbb{R}^{N\times C_1\times \frac{H}{4}\times \frac{W}{4}}$ 
is the output feature of the convolution stem module.
$N$, $C_1$ is the batch size and number of channels. 
Then $X$ is fed to repeated \ourmethod{} blocks, each of which consists of two sub-blocks. 
Specifically, the main components of the first sub-block include \ouratt{} and the BatchNorm module, which we present as
\begin{small}
\begin{equation}
    Y = \mathrm{\ouratt{}}(\mathrm{BN}(X)) + X,
\end{equation}
\end{small}
where $\mathrm{BN}(\cdot)$ denotes a Batch Normalization. 
Details of $\mathrm{\ouratt{}}(\cdot)$ are depicted in Section \ref{Sec:MCA}.
% refers to the module mainly working for mixing token information. 
% It is implemented by various attention mechanism in recent vision transformer models  \cite{vit,refiner,t2t} or spatial MLP in MLP-like models \cite{mlp-mixer, resmlp}. 
% Note that the main function of the token mixer is to propagate token information although some token mixers can also mix channels, like attention. 
The second sub-block consists of two fully-connected layers and a non-linear activation GELU \cite{gelu}.
The output of the MLP module is formulated as follows:
\begin{small}
\begin{equation}
    Z = \mathrm{GELU}(W_1(\mathrm{BN}(Y))W_2 + Y,
\end{equation}
\end{small}
where $W_1 \in \mathbb{R}^{C_i \times eC_i}$ and $W_2 \in \mathbb{R}^{eC_i \times C_i}$ are learnable parameters in fully connected layers.
$e$ is the MLP expansion ratio for the number of channels; 
$C_i$ is the number of channels in the corresponding stage.
% $\sigma(\cdot)$ is a non-linear activation function, such as GELU \cite{gelu} or ReLU \cite{relu} 

% \myPara{Instantiations of MetaFormer} MetaFormer describes a general architecture 
% % a general architecture that is powerful at solving computer vision tasks.  
% with which different models can be obtained immediately by specifying the concrete design of the token mixers. 
% As shown in Figure \ref{fig:first_figure}(a), if the token mixer is specified as attention or spatial MLP, MetaFormer then becomes a transformer or MLP-like model respectively. 

Based on the above \ourmethod{} block, we formulate \ourmethod{}-S and \ourmethod{}-L with different model sizes.
The numbers of channels corresponding to the four stages are identical for both \ourmethod{}-S and \ourmethod{}-L, which are set as 64, 128, 320, and 512.
% \ourmethod{}-S is a smaller-size model with number of channels set as 64, 128, 320, and 512 responding to the four stages. 
Differently, \ourmethod{}-L is larger with more block numbers. Specifically, stages 1, 2, 3, and 4 of \ourmethod{}-S contain $2$, $2$, $6$, $2$ blocks, while that for \ourmethod{}-L are $3$, $3$, $12$, $3$, respectively.
Their detailed architecture designs are shown in Table \ref{tab:model}.

\section{Experiments}

% We evaluate \ourmethod{} on two representative vision tasks, including image classification and semantic segmentation.
% % Ablation studies and visualizations are also presented to thoroughly study the effectiveness of \ourmethod{}.
% % Finally, we show visualization results to provide qualitative analysis.
% All experiments are conducted on 8 NVIDIA A100 GPUs.

\subsection{Image classification}
\label{exp:image-cls}

We conduct image classification experiments on ImageNet-1K \cite{imagenet}.
Each model is trained for 300 epochs with AdamW optimizer, and a total batch size of 1024 on 8 GPUs.
The initial learning rate is set as $1e-3$.
We adopt the cosine decay schedule to adjust the learning rate during training.

\begin{figure}[t]
  \centering
    \includegraphics[width=\linewidth]{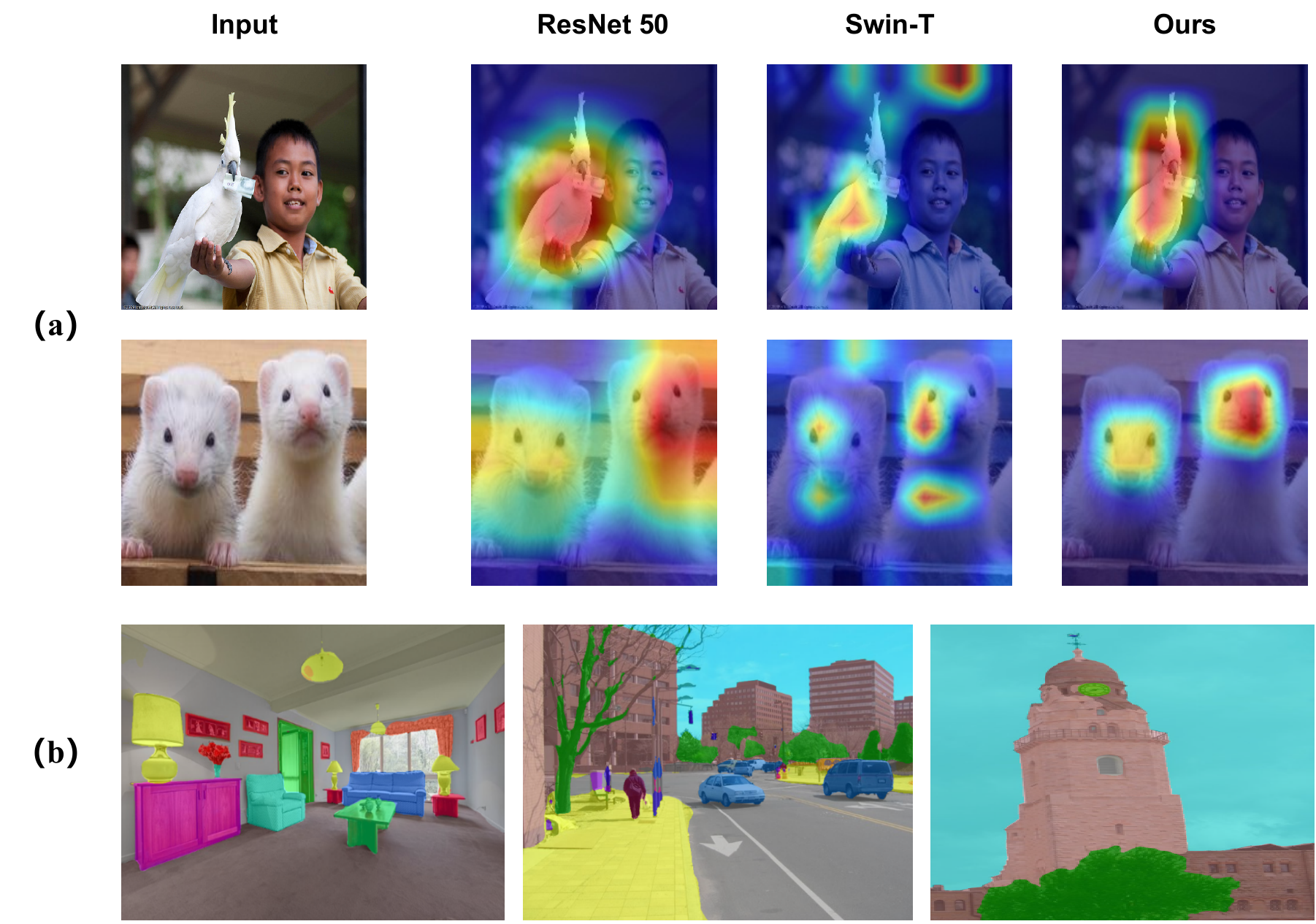}
  \vspace{-3mm}
   \caption{(a) Grad-CAM~\cite{selvaraju2017grad} results generated by Swin-T~\cite{swin}, ResNet50~\cite{resnet} and ours.
   The images are randomly selected from the ImageNet validation dataset.
    Lighter colors in Grad-CAM results refer to stronger attention regions.
   (b) Visualization results of semantic segmentation on ADE20K~\cite{ade20k} with ImageNet pre-trained \ourmethod{}-S as the backbone. 
%   We apply pre-trained \ourmethod{}-S as the backbone, while adopting UperNet~\cite{xiao2018unified} as the framework.
%   Visualization results produced by using Grad-CAM~\cite{selvaraju2017grad} on images selected from the ImageNet validation dataset. 
% % 	All images come from different categories in ImageNet validation set. CAM is produced by using Grad-CAM~\cite{selvaraju2017grad}. 
% 	We present different CAMs generated by  Swin-T~\cite{swin}, ResNet50~\cite{resnet} and ours.
	}
   \label{fig:grad_cam_seg}
     \vspace{-5mm}
\end{figure}

Table \ref{tab:imagenet_cls}~~summarizes the experimental results on ImageNet-1K classification. 
Comparing with recently well-established ViTs like Swin-S \cite{swin} and Focal-S~\cite{yang2021focal}, \ourmethod{} consistently shows better performance. 
Specifically, \ourmethod{}-L surpasses Swin-S, Focal-S by 0.6\%, 0.1\% top-1 accuracy with 9\%, 12\% fewer parameters. 
Comparatively, ConvNeXt \cite{liu2022convnet} is an excellent CNN that learns architecture designs and training schedules from ViTs for better performance.
\ourmethod{}-L outperforms ConvNeXt-S by 0.5\%, while reducing the model size by 10\%.
Moreover, in Fig. \ref{fig:grad_cam_seg} (a), we utilize Grad-CAM \cite{selvaraju2017grad} to localize discriminative regions generated by Swin-T~\cite{liu2021swin}, ResNet50~\cite{resnet} and \ourmethod{}-S.
It can be observed that the highlighted class activation area of \ourmethod{}-S is more accurate.
The experimental results demonstrate the superiority of \ourmethod{}.

\begin{table}[t]\centering
	\caption{Compare with the state-of-the-art methods on the ImageNet validation set. Params means parameter. GFLOPs donates floating point operations.} %Top-1 Acc represents Top-1 accuracy.}
% 	\small
\footnotesize
	\vspace{2mm}
	\resizebox{\linewidth}{!}{
	\begin{tabular}{l|c|c|c}
		\hline
		Method & Params. (M) & GFLOPs & Top-1 Acc (\%)  \\
		\hline
%		ResNet18~\cite{he2016deep} & 11.7 & 1.8 & 69.8  \\
%		PVT-Tiny~\cite{wang2021pyramid} & 13.2 & 1.9 &75.1 \\
%		PoolFormer-S12~\cite{yu2021metaformer} & 11.9  & 2.0 & 77.2 \\  
%		GFNet-XS \cite{rao2021global}  & 16  & 2.9  & 78.6 \\
%		PVTv2-B1~\cite{wang2021pvtv2} & 13.1 & 2.1 & 78.7 \\
%		MixFormer-B3~\cite{chen2022mixformer} & 17 & 1.9 & 81.7 \\
%		VAN-Small~\cite{guo2022visual}  & 13.9 & 2.5 & 81.1 \\
%		\hline
% 		ResNet50~\cite{he2016deep}   &25.6 &4.1 &76.5 \\
% 		ResNeXt50-32x4d~\cite{xie2017aggregated} &25.0 &4.3 &77.6  \\
% 		RegNetY-4G~\cite{regnet} & 21.0 & 4.0 & 80.0 \\
% 		DeiT-Small/16~\cite{touvron2021training}  & 22.1 & 4.6 & 79.8 \\
% 		T2T-ViT$_t$-14~\cite{Yuan_2021_ICCV} & 21.5 & 6.1 & 81.7 \\
		PVT-Small~\cite{pvt}  & 24.5 & 3.8 & 79.8 \\
% 		TNT-S~\cite{tnt} & 23.8 & 5.2 & 81.5 \\
% 		gMLP-S~\cite{liu2021pay} & 20.0 & 4.5 & 79.6 \\
		Swin-T~\cite{swin} & 28.3 & 4.5 & 81.3 \\
% 		PoolFormer-S24~\cite{yu2021metaformer} & 21.4  & 3.6 & 80.3 \\  
		PoolFormer-S36~\cite{yu2022metaformer} & 31.0  & 5.2  &  81.4 \\
% 		ResMLP-24~\cite{resmlp}  & 30.0  & 6.0   & 79.4 \\
		Twins-SVT-S~\cite{chu2021twins} & 24.0 & 2.8 & 81.7 \\
% 		GFNet-S \cite{rao2021global}  & 25.0  & 4.5  & 80.0  \\
% 		PVTv2-B2~\cite{wang2021pvtv2} & 25.4 & 4.0 & 82.0 \\
		Focal-T~\cite{yang2021focal} & 29.1 & 4.9 & 82.2 \\
		ConvNeXt-T~\cite{liu2022convnet} & 28.6 & 4.5 & 82.1 \\  
% 		Shunted-S~\cite{ren2022shunted}  & 22.4 & 4.9  & 82.9 \\
% 		VAN-Base~\cite{guo2022visual}  & 26.6 & 5.0 & \textbf{82.8} \\  
		\ourmethod{}-S   & 26.7   & 5.0  & \textbf{82.8} \\
		\hline
% 		MixFomer-B4~\cite{chen2022mixformer}  & 35.0 & 3.6 & 83.0 \\
% 		ResMLP-24~\cite{touvron2021resmlp} & 30.0 & 6.0 & 79.4 \\
% 		CSWin-S~\cite{dong2022cswin} & 35.0  & 6.9   & 83.6 \\
% 		CvT-21~\cite{wu2021cvt} & 32   & 7.1   & 82.5 \\
% 		T2Tt-19~\cite{t2t}  & 39  & 9.8   & 82.2 \\
% 		Shunted-B~\cite{ren2022shunted}  & 39.6  & 8.1  & 84.0  \\
% 		\hline
% 		VAN-B3~\cite{guo2022visual}  & 45 & 9.0 & 83.9  \\
		PVT-Medium~\cite{pvt}  & 44.2  & 6.7  & 81.2 \\
% 		PVTv2-B3~\cite{wang2021pvtv2}  & 45.2 & 6.9  & 83.2 \\
% 		MixFormer-B5~\cite{chen2022mixformer} & 62  & 6.8  & 83.5 \\
		Focal-S~\cite{yang2021focal}   & 51.1  & 9.1 & 83.5 \\
		Swin-S~\cite{swin}   & 49.6 & 8.7  & 83.0 \\
		ConvNeXt-S~\cite{liu2022convnet}     & 50.0 & 8.7  & 83.1 \\
		\ourmethod{}-L        & 45.0   & 8.7  & \textbf{83.6} \\
		\hline
	\end{tabular}}
	\label{tab:imagenet_cls}
	  	\vspace{-5mm}
\end{table}

\begin{table}[t]
	\centering
	\caption{Results of semantic segmentation on ADE20K~\cite{ade20k} validation set. 
	The pre-trained \ourmethod{}-S is applied as the backbone and plugged in Semantic FPN~\cite{kirillov2019panoptic} and UperNet~\cite{xiao2018unified} frameworks. 
	Note that the difference in the number of parameters between two \ourmethod{}-S is due to the usage of semantic FPN and UperNet.
% FLOPs are calculated with input size of 512 $\times$ 512 for Semantic FPN~\cite{kirillov2019panoptic} and 2,048 $\times$ 512 for UperNet~\cite{xiao2018unified}.
% 	We calculate FLOPs with input size 512 $\times$ 512 for Semantic FPN~\cite{kirillov2019panoptic} and 2,048 $\times$ 512 for UperNet~\cite{xiao2018unified}.
	}
% 	\small
	\footnotesize
	\label{tab:seg}
	\vspace{2mm}
	\resizebox{\linewidth}{!}{
	\begin{tabular}{c|c|c|c}
		\toprule
		\renewcommand{\arraystretch}{0.1}
		Method & Backbone & \#Param (M)  & mIoU (\%)  \\
		\hline
		\multirow{8}{*}{Semantic FPN~\cite{kirillov2019panoptic}} 
%		& ResNet18~\cite{he2016deep} & 16 & 32 & 32.9 \\
%		& PVT-Tiny~\cite{wang2021pyramid} & 17 & 33 & 35.7 \\
%		& PoolFormer-S12~\cite{yu2021metaformer} & 16 & 31 & 37.2 \\
%		& PVTv2-B1~\cite{wang2021pvtv2} & 18 & 34  & 42.5 \\
%		& VAN-Small & 18 & 35 & \textbf{42.9} \\
		
%		\cline{2-5}
% 		& ResNet50~\cite{he2016deep} & 29 & 46 & 36.7\\
% 		& PVT-Small~\cite{wang2021pyramid} & 28 & 45 & 39.8 \\
		& ResNet101~\cite{resnet} & 47.5  & 38.8\\
		& ResNeXt101-32x4d~\cite{xie2017aggregated} & 47.1  & 39.7 \\
		& PoolFormer-S36~\cite{yu2022metaformer} & 34.6   & 42.0 \\
% 		& PVTv2-B2~\cite{wang2021pvtv2} & 29   & 45.2 \\
		& PVT-Medium~\cite{pvt} & 48.0  & 41.6\\
% 		& PoolFormer-S36~\cite{yu2021metaformer} & 35  & 42.0 \\
% 		& PVTv2-B3~\cite{wang2021pvtv2} & 49  & 47.3 \\
% 		& Swin-T~\cite{liu2021swin} & 31.9    & 41.5 \\
		& TwinP-S~\cite{chu2021twins} & 28.4    & 44.3  \\
% 		& Twin-S~\cite{chu2021twins}  & 28.3    & 43.2   \\
		& VAN-B2~\cite{guo2022visual} & 30.0  & 46.7 \\
		& \ourmethod{}-S   & 30.4  & \textbf{47.2} \\
% 		& Shunted-S~\cite{ren2022shunted}  & 26.1  & 183  & 48.2   \\
		\midrule
		
		% DANet~\cite{fu2019dual} & \multirow{4}{*}{ResNet-101~\cite{he2016deep}} & 69 & 1119 & 45.2  \\
		% DLab.v3+~\cite{chen2018encoder} &  & 63 & 1021 & 44.1  \\
%		UperNet~\cite{xiao2018unified} & \multirow{3}{*}{ResNet-101~\cite{he2016deep}} & 86 & 1029 & 44.9  \\
%		OCRNet~\cite{yuan2020object} &  & 56 & 923 & 45.3 \\
%		HamNet~\cite{geng2021attention} & & 69 & 1111 & 46.8  \\
		
%		\hline
		\multirow{5}{*}{UperNet~\cite{xiao2018unified}} 
        % & PoolFormer-M36~\cite{yu2021metaformer}  & 59.8  & 42.4 \\	
% 		& PVT-Large~\cite{wang2021pyramid}  & 65.1 & 42.1 \\
		& ConvNeXt-T~\cite{liu2022convnet}  & 60.0     & 46.7  \\
%		& VAN-Tiny & 32 & 858 & 41.1 \\
% 		& VAN-Small & 44 & 895 & 44.9 \\
% 		& VAN-Base & 57 & 948 &  48.3 \\
        & Swin-T~\cite{swin}  & 60.0  & 46.1 \\
		& TwinP-S~\cite{chu2021twins}   & 54.6   & 46.2  \\
		& Focal-T~\cite{yang2021focal}  & 62.0      & 45.8  \\
% 		& Shunted-S~\cite{ren2022shunted} & 52    & 48.9  \\
        & \ourmethod{}-S    & 56.6  & \textbf{47.6} \\
% 		\cline{2-5}
% 		& Swin-S~\cite{liu2021swin}  & 81 & 1038 & 49.3  \\
% 		& ResNet-101~\cite{he2016deep} & 86 & 1029 & 44.9 \\
		
		\bottomrule
	\end{tabular}}
% 	}
  	\vspace{-4mm}
\end{table}

\subsection{Semantic segmentation}
\label{exp:seg}
\vspace{-3pt}

We evaluate models for the semantic segmentation task on ADE20K~\cite{ade20k}.
mIoU (mean Intersection Over Union) is applied to measure the model performance.
Semantic FPN \cite{kirillov2019panoptic} and UperNet \cite{xiao2018unified} are used as main frameworks to evaluate the \ourmethod{}-S backbone.
Pre-trained weights on ImageNet-1K are utilized to initialize our backbone.
We train each model with AdamW optimizer, a total batch size of 16 on 8 GPUs.
When equipped with Semantic FPN \cite{kirillov2019panoptic}, we use the 40K-iteration training scheme in \cite{pvt, yu2022metaformer}.
The learning rate is set as $2\times10^{-4}$ and decays by polynomial schedule with a power of 0.9.
Comparatively, we adopt the 160K-iteration training scheme in \cite{swin} for UperNet.
Specifically, the learning rate is set as $6\times10^{-5}$ with 1500 iteration warmup and linear learning rate decay.

Table~\ref{tab:seg}~ lists the performance of different backbones using FPN and UperNet. 
Generally, \ourmethod{} consistently outperforms other state-of-the-art backbones.  
When using Semantic FPN for semantic segmentation, \ourmethod{}-S surpasses VAN-B2~\cite{guo2022visual} by 0.5 mIoU with similar computation cost.
Moreover, although the parameter of \ourmethod{}-S is 37\% smaller than that of PVT-Medium~\cite{pvt}, the mIoU is still 5.6 points higher (47.2 vs 41.6).
When applying UperNet as the framework, our model outperforms ConvNeXt-T~\cite{liu2022convnet} by 0.9 mIOU, with about 6\% model size reduction.
Additionally, \ourmethod{}-S is 1.8 mIOU higher than Focal-T~\cite{yang2021focal}  with 9\% decreased parameters.
In Fig. \ref{fig:grad_cam_seg} (b), we present visualization of semantic segmentation results applying UperNet~\cite{xiao2018unified}. 
The results indicate the powerful capacity of \ourmethod{} for the semantic segmentation task.

\subsection{Ablation studies}
\label{sec:ablation}
We conduct ablation studies on each component of \ouratt{} module to shed light on various architecture designs.
\ourmethod{}-S is adopted as the baseline model.
In the w/o multi-scale Conv variant, we only keep a single convolution branch with $7\times7$ kernel size. 
For w/o Expansion rate variant, we set $r$ in the \ouratt{} (see Table \ref{tab.architecture} and Fig. \ref{fig:MCA}) as 1.
To achieve comparable model complexity, we modify block numbers in some variants.
The experimental results in Table \ref{tab_ablation} indicate that all components in \ouratt{} are non-trivial to improve performance.

% \begin{figure}[t]
%   \centering
% %   \includegraphics[width=0.95\linewidth]{figures/PoolFormer_overall_architecture.pdf}
%     \includegraphics[width=0.95\linewidth]{figs/seg_visualize.pdf}
% %   \vspace{-4mm}
%   \caption{Visualization results of semantic segmentation on ADE20K~\cite{ade20k}. 
%   We apply pre-trained \ourmethod{}-S as the backbone, while adopting UperNet~\cite{xiao2018unified} as framework to generate semantic segmentation results.}
%   \label{fig:seg}
% \end{figure}

\begin{table}[t]
	\centering
	\footnotesize
	\caption{Ablation study of different modules in \ouratt{}.
	Block numbers are modified to achieve comparable computation complexity.
		We adopt Semantic FPN \cite{kirillov2019panoptic} as main framework, and use ImageNet pre-trained \ourmethod{}-S variants as backbone for semantic segmentation.
		Parameters and FLOPs are calculated under the ImageNet classification setting.
% 		w/o Attention means we adopt \figref{fig:attention}(b).
	}
	\label{tab_ablation}
	\vspace{2mm}
	\resizebox{\linewidth}{!}{	
\begin{tabular}{c|c|c|c|c|c}
\hline
\multirow{2}{*}{Variants} & \multirow{2}{*}{Blocks} & Params.  & FLOPs & ImageNet  & ADE20k \\
                                      &                         &    (M)                          &    (G)                       & Top-1 Acc  & mIoU   \\ \hline
w/o Expansion rate $r$                   & 2, 2, 12, 2              & 26.9                         & 5.0                       & 82.8          & 46.3       \\
w/o Multi-scale Conv                      & 2, 2, 6, 2                & 26.2                         & 4.9                       & 82.5             & 46.0       \\
w/o Dilation                      &  2, 2, 6, 2                & 26.7                         & 5.0                       & 82.4                 &  46.4      \\
w/o Gating Mechanism                  &  2, 2, 7, 2                & 26.2                         & 5.4                       &  82.7                & 45.7       \\
\ourmethod{}-S                        &  2, 2, 6, 2                & 26.7                         & 5.0                       & \textbf{82.8}         & \textbf{47.2 }  \\ \hline
\end{tabular}}
	\label{tab:ablation}
		\vspace{-4mm}
\end{table}

\section{Conclusion}

In this paper, we have introduced \ourmethod{}, a novel efficient vision backbone that outperforms most similar-sized state-of-the-art models on ImageNet classification and semantic segmentation tasks. 
We have also presented \ouratt{}, the key component of \ourmethod{} which shows much efficiency over the existing self-attention mechanisms.
% \ouratt{}  across FLOPs, throughput and memory consumption. 
Future work may include adapting \ourmethod{} to other tasks such as video processing and object detection, or extending \ouratt{} to a hybrid design with existing excellent CNN or self-attention mechanisms to further enhance performance.

\clearpage

\bibliographystyle{IEEEbib}
\bibliography{mybibfile}

\begin{thebibliography}{10}

\bibitem{vit}
Alexey Dosovitskiy, Lucas Beyer, Alexander Kolesnikov, and Dirk Weissenborn,
\newblock ``An image is worth 16x16 words: Transformers for image recognition
  at scale,''
\newblock in {\em International Conference on Learning Representations}, 2020.

\bibitem{deit}
Hugo Touvron, Matthieu Cord, Alexandre Sablayrolles, and Herv{\'e} J{\'e}gou,
\newblock ``Training data-efficient image transformers \& distillation through
  attention,''
\newblock in {\em International Conference on Machine Learning}. PMLR, 2021,
  pp. 10347--10357.

\bibitem{swin}
Ze~Liu, Yutong Lin, Yue Cao, and Baining Guo,
\newblock ``Swin transformer: Hierarchical vision transformer using shifted
  windows,''
\newblock in {\em Proceedings of the IEEE/CVF International Conference on
  Computer Vision (ICCV)}, October 2021, pp. 10012--10022.

\bibitem{huang2021shuffle}
Zilong Huang, Youcheng Ben, Guozhong Luo, Pei Cheng, Gang Yu, and Bin Fu,
\newblock ``Shuffle transformer: Rethinking spatial shuffle for vision
  transformer,''
\newblock {\em arXiv preprint arXiv:2106.03650}, 2021.

\bibitem{liu2022convnet}
Zhuang Liu, Hanzi Mao, Chao-Yuan Wu, Christoph Feichtenhofer, Trevor Darrell,
  and Saining Xie,
\newblock ``A convnet for the 2020s,''
\newblock {\em arXiv preprint arXiv:2201.03545}, 2022.

\bibitem{resnet}
Kaiming He, Xiangyu Zhang, Shaoqing Ren, and Jian Sun,
\newblock ``Deep residual learning for image recognition,''
\newblock in {\em CVPR}, 2016, pp. 770--778.

\bibitem{ding2022scaling}
Xiaohan Ding, Xiangyu Zhang, Jungong Han, and Guiguang Ding,
\newblock ``Scaling up your kernels to 31x31: Revisiting large kernel design in
  cnns,''
\newblock in {\em CVPR}, 2022, pp. 11963--11975.

\bibitem{han2021connection}
Qi~Han, Zejia Fan, Qi~Dai, Lei Sun, Ming-Ming Cheng, Jiaying Liu, and Jingdong
  Wang,
\newblock ``On the connection between local attention and dynamic depth-wise
  convolution,''
\newblock in {\em International Conference on Learning Representations}, 2021.

\bibitem{regnet}
Ilija Radosavovic, Raj~Prateek Kosaraju, Ross Girshick, Kaiming He, and Piotr
  Doll{\'a}r,
\newblock ``Designing network design spaces,''
\newblock 2020, pp. 10428--10436.

\bibitem{wang2021pvtv2}
Wenhai Wang, Enze Xie, Tong Lu, Ping Luo, and Ling Shao,
\newblock ``Pvtv2: Improved baselines with pyramid vision transformer,''
\newblock {\em arXiv preprint arXiv:2106.13797}, 2021.

\bibitem{tan2019mixconv}
Mingxing Tan and Quoc~V Le,
\newblock ``Mixconv: Mixed depthwise convolutional kernels,''
\newblock {\em arXiv preprint arXiv:1907.09595}, 2019.

\bibitem{gelu}
Dan Hendrycks and Kevin Gimpel,
\newblock ``Gaussian error linear units (gelus),''
\newblock {\em arXiv preprint arXiv:1606.08415}, 2016.

\bibitem{imagenet}
Jia Deng, Wei Dong, Richard Socher, Li-Jia Li, Kai Li, and Li~Fei-Fei,
\newblock ``Imagenet: A large-scale hierarchical image database,''
\newblock in {\em 2009 IEEE conference on computer vision and pattern
  recognition}. Ieee, 2009, pp. 248--255.

\bibitem{yang2021focal}
Jianwei Yang, Chunyuan Li, Pengchuan Zhang, Xiyang Dai, Bin Xiao, Lu~Yuan, and
  Jianfeng Gao,
\newblock ``Focal self-attention for local-global interactions in vision
  transformers,''
\newblock {\em arXiv preprint arXiv:2107.00641}, 2021.

\bibitem{selvaraju2017grad}
Ramprasaath~R Selvaraju, Michael Cogswell, Abhishek Das, Ramakrishna Vedantam,
  Devi Parikh, and Dhruv Batra,
\newblock ``Grad-cam: Visual explanations from deep networks via gradient-based
  localization,''
\newblock 2017, pp. 618--626.

\bibitem{liu2021swin}
Ze~Liu, Yutong Lin, Yue Cao, Stephen Lin, and Baining Guo,
\newblock ``Swin transformer: Hierarchical vision transformer using shifted
  windows,''
\newblock 2021.

\bibitem{ade20k}
Bolei Zhou, Hang Zhao, and Antonio Torralba,
\newblock ``Scene parsing through ade20k dataset,''
\newblock in {\em CVPR}, 2017, pp. 633--641.

\bibitem{pvt}
Wenhai Wang, Enze Xie, Ping Song, and Ling Shao,
\newblock ``Pyramid vision transformer: A versatile backbone for dense
  prediction without convolutions,''
\newblock in {\em ICCV}, October 2021, pp. 568--578.

\bibitem{yu2022metaformer}
Weihao Yu, Mi~Luo, and Shuicheng Yan,
\newblock ``Metaformer is actually what you need for vision,''
\newblock in {\em CVPR}, 2022, pp. 10819--10829.

\bibitem{chu2021twins}
Xiangxiang Chu, Zhi Tian, Yuqing Wang, Bo~Zhang, Haibing Ren, Xiaolin Wei,
  Huaxia Xia, and Chunhua Shen,
\newblock ``Twins: Revisiting the design of spatial attention in vision
  transformers,''
\newblock vol. 34, 2021.

\bibitem{kirillov2019panoptic}
Alexander Kirillov, Ross Girshick, Kaiming He, and Piotr Doll{\'a}r,
\newblock ``Panoptic feature pyramid networks,''
\newblock 2019, pp. 6399--6408.

\bibitem{xiao2018unified}
Tete Xiao, Yingcheng Liu, Bolei Zhou, Yuning Jiang, and Jian Sun,
\newblock ``Unified perceptual parsing for scene understanding,''
\newblock 2018, pp. 418--434.

\bibitem{xie2017aggregated}
Saining Xie, Ross Girshick, Piotr Doll{\'a}r, Zhuowen Tu, and Kaiming He,
\newblock ``Aggregated residual transformations for deep neural networks,''
\newblock in {\em CVPR}, 2017, pp. 1492--1500.

\bibitem{guo2022visual}
Meng-Hao Guo, Cheng-Ze Lu, Zheng-Ning Liu, Ming-Ming Cheng, and Shi-Min Hu,
\newblock ``Visual attention network,''
\newblock {\em arXiv preprint arXiv:2202.09741}, 2022.

\end{thebibliography}

\end{document}